\documentclass[10pt, conference]{IEEEtran} 
\usepackage{graphicx} 
\usepackage{caption} 
\usepackage{algorithm}
\usepackage{cite}

\usepackage[caption=false,font=footnotesize]{subfig}
\usepackage{tabularx, booktabs}
\usepackage{algpseudocode}
\usepackage{algorithmicx}
\usepackage{amsmath}
\usepackage{amssymb}
\usepackage{balance}
\usepackage{cleveref}
\usepackage{xcolor}

\usepackage{color}
\definecolor{Orange}{rgb}{0.9,0.5,0}
\definecolor{Magenta}{rgb}{0.8, 0.1, 0.6}
\definecolor{BrightCerulean}{rgb}{0.11, 0.67, 0.84}

\def\etal{\emph{et al.}}
\def\ie{\emph{i.e.}}
\def\eg{\emph{e.g.}}

\title{Multimodal Federated Learning on IoT Data}

\author{\IEEEauthorblockN{Yuchen Zhao}
\IEEEauthorblockA{UK Dementia Research Institute \\
Imperial College London\\
yuchen.zhao19@imperial.ac.uk}
\and
\IEEEauthorblockN{Payam Barnaghi}
\IEEEauthorblockA{UK Dementia Research Institute \\
Imperial College London\\
p.barnaghi@imperial.ac.uk}
\and
\IEEEauthorblockN{Hamed Haddadi}
\IEEEauthorblockA{UK Dementia Research Institute \\
Imperial College London\\
h.haddadi@imperial.ac.uk}
}

\begin{document}
\maketitle
\begin{abstract}
    Federated learning is proposed as an alternative to centralized machine
    learning since its client-server structure provides better privacy
    protection and scalability in real-world applications. In many applications,
    such as smart homes with Internet-of-Things (IoT) devices, local data on
    clients are generated from different modalities such as sensory, visual, and
    audio data. Existing federated learning systems only work on local data from
    a single modality, which limits the scalability of the systems. 
    
    In this paper, we propose a multimodal and semi-supervised federated
    learning framework that trains autoencoders to extract shared or correlated
    representations from different local data modalities on clients. In
    addition, we propose a multimodal FedAvg algorithm to aggregate local
    autoencoders trained on different data modalities. We use the learned global
    autoencoder for a downstream classification task with the help of auxiliary
    labelled data on the server. We empirically evaluate our framework on
    different modalities including sensory data, depth camera videos, and RGB
    camera videos. Our experimental results demonstrate that introducing data
    from multiple modalities into federated learning can improve its
    classification performance. In addition, we can use labelled data from only
    one modality for supervised learning on the server and apply the learned
    model to testing data from other modalities to achieve decent $F_1$ scores
    (\eg, with the best performance being higher than $60\%$), especially when
    combining contributions from both unimodal clients and multimodal clients.
    \end{abstract}

\section{Introduction}
\label{sec:intro}
In recent years, we have witnessed a rapid growth in personal data generated
from many different aspects in people's daily lives, such as mobile devices and
IoT devices. Powered by the enormous amount of personal data, machine-learning
(ML) techniques, especially Deep Neural Networks (DNN), have shown great
capabilities of conducting complex tasks such as image recognition, natural
language processing, human activity recognition, and so forth. Traditionally, ML
systems are centralized and need to collect and store personal data on a server
to train DNN models, which causes privacy issues. The long-debated privacy
issues in centralized ML systems have motivated researchers to design and
implement machine learning in decentralized fashions. Federated learning
(FL)~\cite{McMahan2016}, which allows different parties to jointly train DNN
models without releasing their local data, is a system paradigm that has gained
much popularity in both research communities and real-world ML applications.

In FL systems, DNN models are trained on clients at the edge of networks instead
of on servers in the cloud. This makes FL systems specifically suitable for
privacy sensitive applications such as smart
home~\cite{8844592,9013081,Zhao2020:edgesys} based on IoT technologies. For
example, Wu~\etal~\cite{Wu2020} propose an FL framework that uses
personalization to address the device, statistical and model heterogeneity
issues in IoT environments. Pang~\etal~\cite{Pang2021} propose an FL framework
using reinforcement learning to adjust the model aggregation strategy on models
trained with IoT data. As a distributed system paradigm, FL provides a feasible
and scalable solution for realizing ML on resource-constrained IoT
devices~\cite{Imteaj2021}.

IoT applications often deploy different types of sensors or devices that
generate data from different modalities (\eg, sensory, visual, and
audio)~\cite{Brunete2021}. For example, in one smart home, activities of a
person can be recorded by body sensors in a smartwatch worn by the person, and
also by a video camera in the room at the same time. Meanwhile, for smart homes
with different device setups, some of them may have multimodal local data (\ie,
\emph{multimodal clients}) while the others may have unimodal local data (\ie,
\emph{unimodal clients}).  One way to apply FL to these IoT applications is to
implement individual services for different modalities. However, many
centralized ML systems~\cite{Ngiam2011, Wang2015, Ofli2013, Radu2018, Xing2018}
have shown that combining data from different modalities can improve their
performance. Therefore, it is necessary to design and implement FL systems in a
way that supports multimodal IoT data and different device setups.

To work on multimodal data, one approach in existing FL systems uses data
fusion~\cite{Liang2020} to mix representations from different modalities before
a final decision layer into a new representation space. This requires all the
data (\ie, training and testing) in the system to be aligned multimodal data,
which means that all the clients need to have data from all modalities in the
system. In addition, the labelled data in the system also need to be from all
modalities, in order to support supervised learning on the new representation
space. This does not work on systems with unimodal clients and increases the
complexity of data annotation. Another approach~\cite{Liu2020} extracts
representations from different modalities locally and requires the clients to
send the representations to the server in order to align different modalities.
This may break the privacy guarantee provided by FL since the representations
can be used to recover local data, especially when the server has taken part in
the training of the model that extracts the representations. Allowing FL to work
on clients with arbitrary data modalities (\ie, unimodal or multimodal) and with
labelled data that come from single modalities, however, still remains a
challenge.

In this paper, we propose a multimodal FL framework that takes advantage of
aligned multimodal data on clients. Although acquiring alignment
information for multimodal data across different clients is challenging, our
assumption is that data from different modalities (\eg, sensory data and visual
data) on a \emph{multimodal client} inherently have some alignment information
(\eg, through synchronized local timestamps of sensory data samples and video
frames on that client), based on which we can train models to extract multimodal
representations from the data. We utilize multimodal
autoencoders~\cite{Ngiam2011, Wang2015} to encode the data into shared or
correlated hidden representations. To enable the server in our framework to
aggregate trained local autoencoders into a global autoencoder, we propose a
multimodal version of the FedAvg algorithm~\cite{McMahan2016} that can combine
local models trained on data from both unimodal and multimodal clients.

As it is difficult to have adequate labels on clients in real-world FL
systems~\cite{jeong2020federated,Liang2020}, we focus on semi-supervised
scenarios wherein local data on the clients are unlabelled and the server has an
auxiliary labelled dataset. We use the global
autoencoder and the auxiliary labelled dataset on the server to train a
classifier for activity recognition
tasks~\cite{vanBerlo2020:edgesys,Zhao2021semisupervised} and evaluate its
performance on a variety of multimodal datasets (\eg, sensory and visual).
Compared with existing FL systems~\cite{Liang2020, Liu2020}, our proposed
framework does not share representations of local data to the server.
Additionally, instead of requiring the clients and the server to have aligned
data from all modalities, our framework conducts local training on both
multimodal and unimodal clients, and only needs unimodal labelled data on the
server. Our experimental results indicate that our proposed framework can
improve the classification performance ($F_1$ score) of FL systems in comparison
to unimodal FL, and allows us to use unimodal labelled data to train models that
can be applied to multimodal testing data.

We make the following contributions in this paper:
\begin{itemize}
    \item We propose a multimodal FL framework that works on data from different
    modalities and clients with different device setups, and a multimodal
    FedAvg algorithm.
    \item Complementing the existing knowledge on the benefit of using
    multimodal data in centralized ML, we find that introducing data from more
    modalities into FL also leads to better classification performance.
    \item We show that classifiers trained on labelled data on the server from
    one modality can achieve decent classification $F_1$ scores on testing data from
    other modalities.
    \item We show that combining contributions from both unimodal and multimodal
    clients can further improve the classification $F_1$ scores.
\end{itemize}

\section{Related work}
\label{sec:related}
\subsection{Federated learning}
McMahan~\etal~\cite{McMahan2016} propose federated learning (FL) as an
alternative system paradigm to centralized ML. In an FL system, a server acts as
an coordinator to select clients and to send a global DNN model to the clients. The
clients use their own data to locally train the model and then send the
resulting models back to the server, on which these models are aggregated into a
new global model. The system repeats this process for a number of rounds until
the performance of the global model on a given task converges. The privacy of
the clients' data is protected since the data are never shared with others.
Given its decentralized feature, FL is especially suitable for edge
computing~\cite{Shi2016, Chen2019}, which moves computation to the place where
data are generated.

Canonical FL systems focus on supervised learning that requires all local data
on FL clients to be labelled. In edge computing, data generated from IoT devices
can only be accessed by the data subjects, since FL clients do not share data to
third parties. These data subjects (\ie, end users of an FL system) may not have
time or abilities to annotate their data with labels of a given task, especially
when the task requires expert knowledge (\eg, labelling timer-series sensory
data with clinical knowledge). Therefore, one key challenge of deploying FL in
real-world IoT environments is the lack of labelled data on clients for local
training. In order to address this issue, recent research in FL has been
focusing on unsupervised and semi-supervised FL frameworks through data
augmentation ~\cite{jeong2020federated,liu2020rc, zhang2020benchmarking,
long2020fedsemi, zhang2021106679, kang2020fedmvt, wang2020graphfl,
yang2020federated, Saeed2019, Saeed2021} to generate pseudo labels for local
data, or through unsupervised learning to extract hidden representations from
unlabelled local data~\cite{vanBerlo2020:edgesys, Zhao2021semisupervised}. For
example, van Berlo~\etal~\cite{vanBerlo2020:edgesys} propose to learn hidden
representations through convolutional autoencoders from unlabelled local data on
FL clients. Their results show that the learned representations can empower
downstream tasks such as classifications.
Zhao~\etal~\cite{Zhao2021semisupervised} propose a semi-supervised FL framework
for human activity recognition and compared the performance of different
autoencoders. Their framework shows better performance than data augmentation
schemes do. Our work in this paper follows the path of the latter category.
Compared with the existing research, we enable semi-supervised FL to learn from
multiple data modalities.

\subsection{Heterogeneity in federated learning}
Heterogeneity is one of the most challenging issues~\cite{Kairouz2019,
Li2020:SPM} in FL because models are locally trained on clients. Different
clients may vary in terms of computational capabilities, model structures,
distribution of data, or distribution of features. Among all these issues, the
heterogeneity in distribution of data (\ie, non-IID local data) has attracted
most research efforts~\cite{Smith2017,Zhao2018,Li2019,Chen2020}.
Smith~\etal~\cite{Smith2017} apply multi-task learning to addressing the issue
of training on non-IID data in FL. Instead of training one global model for all
clients, they treat each client as a different task and train separate models
for them. Similarly, Li~\etal~\cite{Li2019} extend federated multi-task
learning to an online fashion and allow new clients to join the system. To
address the heterogeneity in the distribution of features when shifting FL from
one domain to another, Chen~\etal~\cite{Chen2020} propose to use transfer
learning to align the features in lower-stream layers (\eg, fully connected
layers before final output layers). In order to learn from heterogeneous models
(\ie, DNN models with different structures), Lin~\etal~\cite{lin2020ensemble}
propose to use knowledge distillation~\cite{hinton2015distilling} to train
global models of FL based on the output probability distribution from local
models, instead of directly averaging the parameters of them. Existing research,
however, neglected the heterogeneity in data modalities in FL, which is
commonplace in many scenarios such as edge computing, IoT environments, and
mobile computing.

The recent study by Liu~\etal~\cite{Liu2020} applies FL on data from two
modalities (\ie, images and texts) and treats each modality individually, which is
the same as running two individual FL instances. In the study, to align
the two modalities on a server, representations of local data need to be
uploaded to the server. This breaks the privacy guarantee of FL because the
server has the global model that generates the representations from raw data and
could recover the raw data if it has those representations. The framework
proposed by Liang~\etal~\cite{Liang2020} can work on multimodal data only
when the clients' local data, the server's labelled data, and testing data
are all aligned data from both modalities. Instead of aligning the
representations from different modalities, it conducts early fusion (\ie,
element-wise multiplication) on the representations. Thus unimodal data cannot
contribute to the local training and the trained model cannot be used on
unimodal data. Compared to the existing work, we use the alignment information
in local data to learn to extract shared or correlated hidden representations
from multiple modalities. Our scheme does not require sending representations of
local data to the server, which contradicts the motivation of using FL. In
addition, it allows models to be trained and used on unimodal data.

\subsection{Multimodal deep learning}
When training deep learning models for a certain task, the used data can be
generated from a variety of modalities (\eg, recognizing human activities from
IoT sensory data or videos). In order to utilize these data, multimodal deep
learning has attracted much attention from researchers.
Ngiam~\etal~\cite{Ngiam2011} propose to use deep autoencoders~\cite{Baldi2011}
to learn multimodal representations from audio and visual data. The alignment
between the two modalities is done by reconstructing the output for both
modalities from the hidden representation generated by either modality.
Wang~\etal~\cite{Wang2015} compare different multimodal representation learning
techniques and propose to combine both deep canonical correlation
analysis~\cite{Andrew2013} and autoencoders to map data from different
modalities into highly correlated representations instead of one common
representation. These techniques have demonstrated that data from different
modalities can complement each other when learning representations and improve
the overall performance of an ML system. Many applications such as audio-visual
speech recognition~\cite{Ngiam2011}, activity and context
recognition~\cite{Radu2018, Xing2018}, and textual description generation for
images~\cite{Karpathy_2015_CVPR}, have been implemented based on multimodal deep
learning. The recent survey by Baltru\v{s}aitis~\etal~\cite{Baltrusaitis2019}
provides a detailed analysis and taxonomy of multimodal deep learning. In this
paper, we apply multimodal representation learning to FL to address the
heterogeneity issue in local data modalities.

\section{Methodology}
Our goal is to enable FL to work on clients that have different local data
modalities. We first introduce the overall design of our framework. We then
describe the key techniques that we use to extract representations from
multimodal data and the algorithms that we designed to aggregate local models
trained on both unimodal and multimodal clients. 

\label{sec:method}
\subsection{Framework overview}
\begin{figure*}[!t]
    \centering
   \subfloat[][Canonical federated learning]{\includegraphics[width=.5\linewidth]{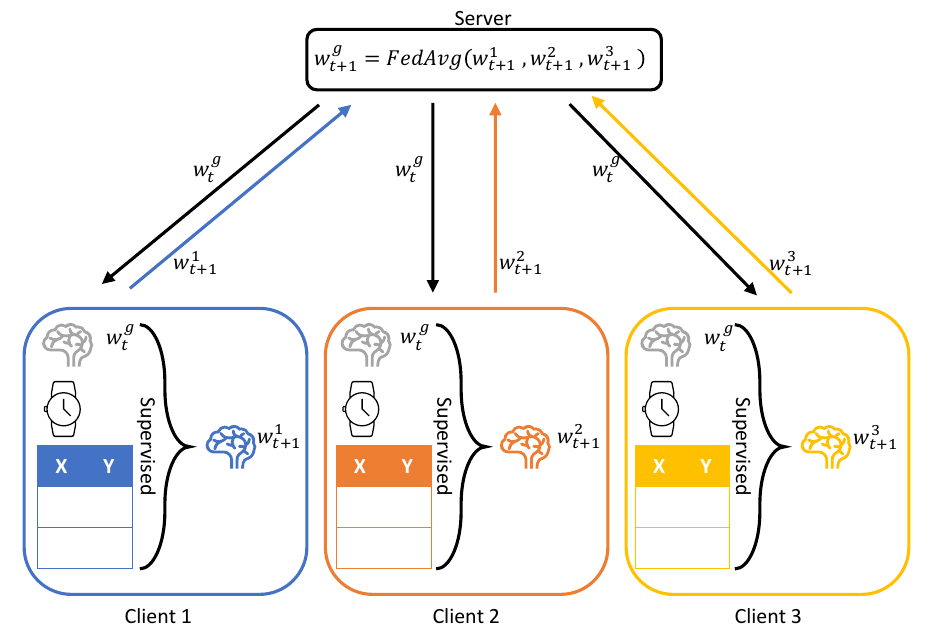}\label{sub_fig:fl}}
   \subfloat[][Multimodal federated learning]{\includegraphics[width=.5\linewidth]{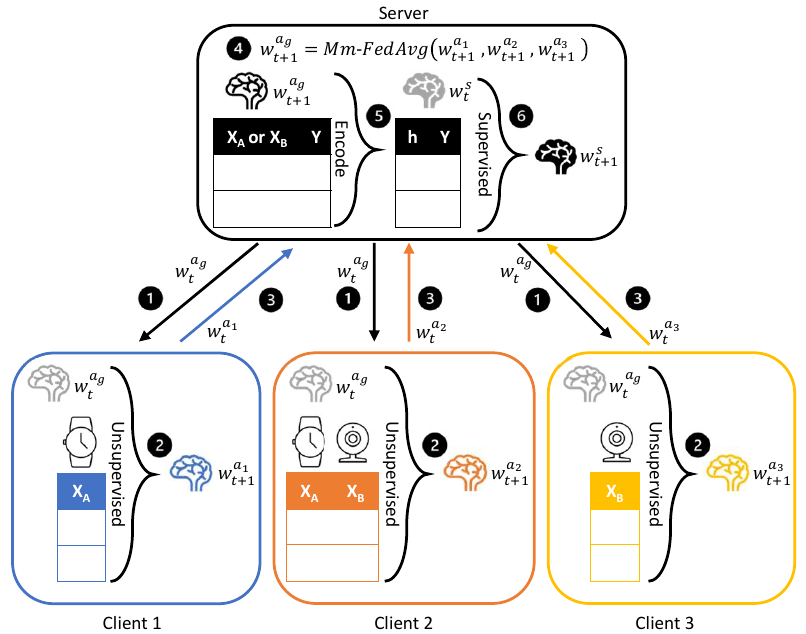}\label{sub_fig:mm-fl}}
   \caption{In canonical federated learning~\ref{sub@sub_fig:fl}, during round
   $t$, a server sends a global model $w^{g}_{t}$ to selected clients that have
   data from the same modality. Client $k$ conducts supervised learning to
   generate a local model $w^{k}_{t+1}$. Local models are aggregated on the
   server by using the FedAvg algorithm. In multimodal federated
   learning~\ref{sub@sub_fig:mm-fl}, a server sends a global model
   $w^{a_{g}}_{t}$ to selected clients to learn to extract multimodal
   representations (Sec.~\ref{subsec:ae}) on unlabelled local data. The server
   uses multimodal FedAvg (Sec.~\ref{subsec:mm-fedavg}) to aggregate local
   models into a new global model $w^{a_{g}}_{t+1}$ and uses it to encode a
   labelled dataset (modality $A$ or $B$) to a labelled representation dataset
   $(h, Y)$. A classifier $w^{s}_{t+1}$ is then trained on $(h, Y)$, which can
   be used by all clients.} 
   \label{fig:overview}
\end{figure*}

A canonical FL system, as shown in Fig.~\ref{sub_fig:fl}, only works on clients that
have local data from the same modality and requires the data to be labelled for
supervised learning.

We propose an FL framework wherein clients' unlabelled local data can be from
either one single modality or multiple modalities. In our framework, as shown in
Fig.~\ref{sub_fig:mm-fl}, unimodal clients (\eg, Clients 1 and 3) only deploy one
type of devices due to reasons such as budget or privacy. Multimodal clients
(\eg, Client 2) deploy both types of devices and thus have multimodal local
data. On a multimodal client, we assume that there is alignment information
between the data from two modalities, based on which we can align the hidden
representations of two modalities. For example, a person's activity can be
captured by the accelerometers in a smartwatch and by an IP camera in the room
at the same time. A record of video call contains both the visual information
and audio information of a speech. This kind of matching information is the key
to align the hidden representations of multimodal data since they describe the
same underlying activities or events.

To address the lack of labelled data in FL systems using IoT devices, similar to
existing semi-supervised FL frameworks~\cite{vanBerlo2020:edgesys,
Zhao2021semisupervised}, on clients we assume that no labelled local data are
available. Thus we learn to extract hidden representations from unlabelled data.
On multimodal clients, we train local models to extract shared or correlated
representations between different modalities since we have aligned pairs of
multimodal data. On unimodal clients, we train models to extract representations
from one single modality. Local models from both types of clients are sent to
the server and are aggregated into a global model by using a multimodal version
of the FedAvg algorithm~\cite{McMahan2016}. The server uses the global model to
encode a labelled dataset from either modality into a labelled representation
dataset, based on which a classifier is trained through supervised learning. We
believe that, as the service provider, the server can provide such an auxiliary
dataset with labels that requires expert knowledge about the task of the
service. For example, in many existing human activity datasets,
labelling activities with sensory data can be done through controlled
laboratory trials with the assistance from video cameras and pre-defined trial
scripts~\cite{Chavarriaga2013}. The clients receive both the global model and
the classifier from the server during each communication round and can use them
on their local data for classifications. Alg.~\ref{alg:mm-fl} describes the the
process of multimodal federated learning.

\begin{algorithm}
    \caption{Multimodal Federated Learning}
    \label{alg:mm-fl}
    \begin{algorithmic}[1]
        \Require $K$: number of clients; $C$: fraction of clients to choose; $D=(X,Y)$:
        labelled dataset from either modality (A or B) 
        \State initializes $w^{a_{g}}_{0}$, $w^{s}_{0}$ at $t=0$
        \ForAll {communication round $t$}
            \State $S_{t}\gets$ randomly selected $K \cdot C$ clients
            \State $W_{t} \gets \emptyset$
            \ForAll {client $k \in S_{t}$}
                \State $w_{t+1}^{a_{k}} \gets \textbf{Multimodal Local Training}(k,w^{a_{g}}_{t})$ \Comment{on client $k$}
                \State $W_{t} \gets W_{t} \cup w_{t+1}^{a_{k}}$
            \EndFor
            \State $w^{a_{g}}_{t+1} \gets \textbf{Multimodal FedAvg}(W_{t})$ \Comment{on the server}
            \State $h \gets w^{a_{g}}_{t+1}.encoder(X)$ \Comment{using the encoder for the modality of $X$}
            \State $D^\prime_{t} \gets (h,Y)$
            \State $w^{s}_{t+1} \gets \textbf{Cloud Training}(D^\prime_{t},w^{s}_{t})$ \Comment{on the server}
        \EndFor
    \end{algorithmic}
\end{algorithm}

\subsection{Learning to extract representations}
\label{subsec:ae}
The key part of the local training in our proposed framework is how to learn
representations from unlabelled unimodal data or multimodal data. We first
introduce canonical autoencoders, which we train to extract hidden
representations from unimodal data. Then we introduce two types of multimodal
autoencoders, which learn to extract shared and correlated hidden
representations from different modalities.

\begin{figure}[t]
    \centering
    \includegraphics[width=.65\linewidth]{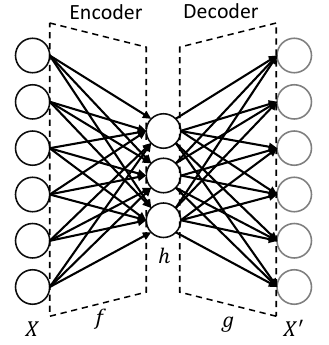}   
    \caption{A simple autoencoder structure. An encoder $f$ maps input data $X$
    into a hidden representation $h$. A decoder $g$ maps $h$ into a
    reconstruction $X^\prime$.}
    \label{fig:ae}
\end{figure}

\begin{figure*}[t]
    \centering
   \subfloat[][Split autoencoder]{\includegraphics[width=.65\linewidth]{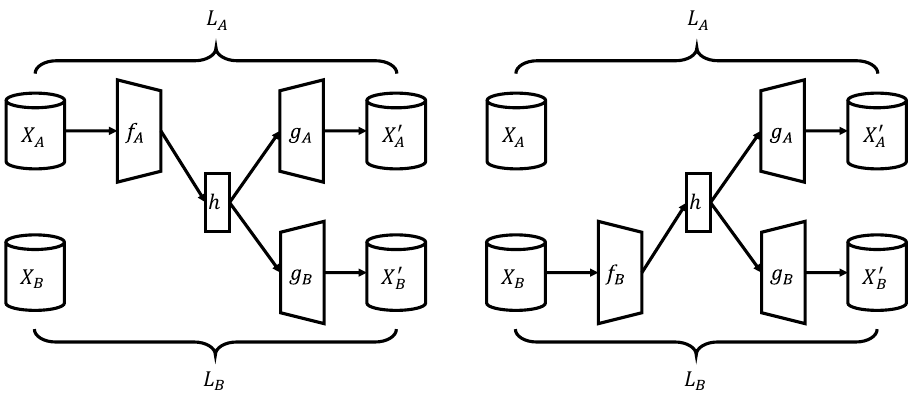}\label{sub_fig:split-ae}}
   \subfloat[][Canonically correlated autoencoder]{\includegraphics[width=.32\linewidth]{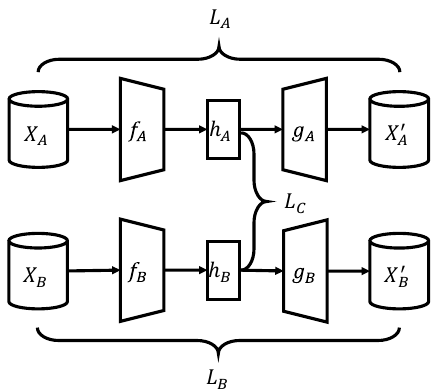}\label{sub_fig:dccae}}
    \caption{In split autoencoders~\ref{sub@sub_fig:split-ae}, for aligned input
    $(X_{A}, X_{B})$ from two modalities, data from one modality are input into
    its encoder to generate an $h$, which is then used to reconstruct the data
    for both modalities through two decoders. Each single modality has a loss
    function (\ie, $L_{A}$ and $L_{B}$) and the overall objective of training is
    to minimize $L_{A}+L_{B}$. In a canonically correlated
    autoencoder~\ref{sub@sub_fig:dccae}, data from both modalities are input
    into their encoders to generate two representations. Two parameter matrices
    are used to maximize the canonical correlation between the paired
    representations $h_{A}$ and $h_{B}$. The overall objective of the training
    is to minimize $\lambda(L_{A}+L_{B})+L_{C}$, where $\lambda$ is a trade-off
    parameter and $L_{C}$ is the negative value of the canonical correlation.}
    \label{fig:models}
\end{figure*}

\subsubsection{Autoencoders}
Autoencoders~\cite{Baldi2011} are one of the most commonly used DNNs in
unsupervised ML. A typical autoencoder, as shown in Fig.~\ref{fig:ae}, has two
building blocks, which are an \emph{encoder} ($f$) and a \emph{decoder} ($g$).
The encoder maps unlabelled data ($X$) into a hidden representation ($h$). The
decoder tries to generate a reconstruction ($X^\prime$) of the input data from
the representation. When training an autoencoder, the objective is to minimize
the difference between $X$ and $X^\prime$, which is measured by a loss function
$L(X,X^\prime)$, such as the mean squared error (MSE). The assumption is that if
the reconstruction error is small, then it means that the hidden representation
contains the most useful information in the original input. Therefore,
minimizing the error will make the encoder to learn to extract such useful
information.

\subsubsection{Split autoencoders}
Canonical autoencoders only work on data from the same modality. In order to
extract shared representations from multimodal data,
Ngiam~\etal~\cite{Ngiam2011} propose a split autoencoder (SplitAE) that takes
input data from one modality and encode the data into a shared $h$ for two
modalities. With the shared $h$, two decoders are used to generate the
reconstructions for two modalities. Fig.~\ref{sub_fig:split-ae} shows the
structures of SplitAEs for two data modalities. The premise is that the data
from two modalities have to be matching pairs, which means that they present the
same underlying activities or events. Since the encoders for both modalities aim
to extract hidden representations, we want the representations to be not only
specific to an individual modality. Instead, we hope that the extracted
representations from both encoders can reflect the general nature of the
activities or events in question.

For modalities $A$ and $B$, given a pair of matching samples $(X_{A},X_{B})$
(\eg, accelerometer data and video data of the same activity), the SplitAE
$(f_{A}, g_{A}, g_{B})$ for input modality $A$ is:

\begin{equation}
\label{eq:split-ae}
\underset{f_{A}, g_{A}, g_{B}}{\arg\min}\ L_{A}(X_{A}, X_{A}^\prime)+L_{B}(X_{B}, X_{B}^\prime)
\end{equation}

$X^\prime_{A}$ and $X^\prime_{B}$ are the reconstructions for two modalities.
$L_{A}$ and $L_{B}$ are the loss functions for two modalities, respectively. By
minimizing the compound loss in Eq.~\ref{eq:split-ae}, the learned encoder
$f_{A}$ will extract representations that are useful for both modalities.
Similarly, for input modality $B$, its SplitAE is $(f_{B}, g_{A}, g_{B})$.

\subsubsection{Deep canonically correlated autoencoders}
In order to combine deep canonical correlation analysis~\cite{Andrew2013} and
autoencoders together, Wang~\etal~\cite{Wang2015} propose a deep canonically
correlated autoencoder (DCCAE). Instead of mapping multimodal data into shared
representations, DCCAE keeps an individual autoencoder for each modality and
tries to maximize the canonical correlation between the hidden representations
from two modalities. Fig.~\ref{sub_fig:dccae} shows the structure of a DCCAE for
two modalities.

For modalities $A$ and $B$, given aligned input
$(X_{A},X_{B})$, the DCCAE $(f_{A}, g_{A}, f_{B}, g_{B})$ is:

\begin{align}
   \underset{f_{A}, g_{A}, f_{B}, g_{B}, U, V}{\arg\min}\ \lambda(L_{A}+L_{B})+L_{C} \\
   L_{C} = -\text{tr}(U^\intercal f_{A}(X_{A})f_{B}(X_{B})^\intercal V)
\end{align}

Parameter matrices $U$ and $V$ are canonical correlation analysis directions.
Similarly to SplitAE, one of the objectives of DCCAE is to minimize the
reconstruction losses. In addition, it uses another objective to increase the
canonical correlation between the generated representations from two modalities
(\ie, minimizing its negative value $L_{C}$). The two objectives are balanced by
a parameter $\lambda$. By this means, DCCAE maps multimodal data into correlated
representations rather than shared representations.

\begin{figure*}[t]
    \centering
    \includegraphics[width=.6\linewidth]{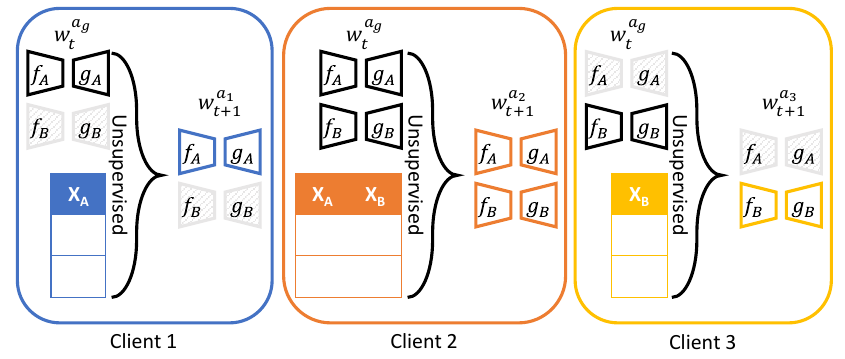}
    \caption{Multimodal local training. Clients only update the $f$ and
    $g$ that are related to the modalities of their data.}
    \label{fig:mm-training}
\end{figure*}

\begin{figure}[t]
    \centering
    \includegraphics[width=.6\linewidth]{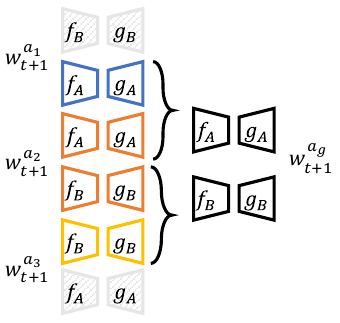}
    \caption{Multimodal FedAvg on the server. Only the updated
    parts of each local model will be aggregated.}
    \label{fig:mm-fedavg}
\end{figure}

\subsection{Multimodal federated averaging}
\label{subsec:mm-fedavg}
During each round $t$, the server sends a global multimodal autoencoder
$w^{a_{g}}_{t}$ to selected clients. A selected client is either unimodal or
multimodal and the local training on $w^{a_{g}}_{t}$ depends on the modality of
data on the client. As shown in Fig.~\ref{fig:mm-training}, a multimodal
client (\eg, Client 2) locally updates the encoders and decoders for both
modalities. A unimodal client (\eg, Client 1 or 3) only updates the encoder and
decoder for its data modality through standard autoencoder training. The encoder
and decoder for the other modality will be frozen during the local training.

We propose a multimodal FedAvg (Mm-FedAvg) algorithm to aggregate autoencoders
received from both unimodal clients and multimodal clients.
Fig.~\ref{fig:mm-fedavg} shows which parts of different local autoencoders
are used when generating a new global model. Given a global multimodal
autoencoder $w^{a_{g}}_{t}$ at round $t$ represented as $(f_{A}, g_{A},
f_{B},g_{B})_{t}$, $(f_{A},g_{A})_{t}$ is the encoder and decoder for modality
$A$. Similarly, a local multimodal autoencoder updated by client $k$ is
$w^{a_{k}}_{t}$ and the client's modality $m_{k}$ is one of $A$, $B$ and $AB$.
The Mm-FedAvg algorithm is shown in Alg.~\ref{alg:mm-avg}.

\begin{algorithm}
  \caption{Multimodal FedAvg (Mm-FedAvg)}
  \label{alg:mm-avg}
  \begin{algorithmic}[1]
      \Require $W_{t}$: local multimodal autoencoders at round $t$; $\alpha$: multimodal weight parameter; $n^{k}$: number of samples on client $k$; $m^{k}$: data modality of client $k$;
      \State $W_{t}^A \gets \{w^{a_k}|w^{a_k} \in W_{t} \land m^{k}=A\}$
      \State $W_{t}^B \gets \{w^{a_k}|w^{a_k} \in W_{t} \land m^{k}=B\}$
      \State $W_{t}^{AB} \gets \{w^{a_k}|w^{a_k} \in W_{t} \land m^{k}=AB\}$
      \State $n_{A} \gets \sum_{w^{a_k} \in W_{t}^A} n^{k} + \alpha\sum_{w^{a_k} \in W_{t}^{AB}} n^{k}$
      \State $n_{B} \gets \sum_{w^{a_k} \in W_{t}^B} n^{k} + \alpha\sum_{w^{a_k} \in W_{t}^{AB}} n^{k}$
      \State $(f_{A}, g_{A}) \gets \sum_{w^{a_k} \in W_{t}^A} \frac{n^{k}}{n_{A}}(f_{A},g_{A})^{k} + \alpha\sum_{w^{a_k} \in W_{t}^{AB}} \frac{n^{k}}{n_{A}}(f_{A},g_{A})^{k}$
      \State $(f_{B}, g_{B}) \gets \sum_{w^{a_k} \in W_{t}^B} \frac{n^{k}}{n_{B}}(f_{B},g_{B})^{k} + \alpha\sum_{w^{a_k} \in W_{t}^{AB}} \frac{n^{k}}{n_{B}}(f_{B},g_{B})^{k}$
      \State $w^{a_{g}}_{t+1} \gets (f_{A},g_{A}, f_{B}, g_{B})$
  \end{algorithmic}
\end{algorithm}

When aggregating local models from multimodal clients and unimodal clients, the
contribution from multimodal clients is controlled by a weight parameter
$\alpha$. Increasing $\alpha$ can give more weights to multimodal clients
because they play a key role in aligning two modalities, which helps unimodal
clients benefit from the data from another modality.

\section{Evaluation}
We evaluate our proposed framework on different multimodal datasets including
sensory data, depth camera data, and RGB camera data through simulations. The
research questions that we want to answer are as follows:

\begin{itemize}
    \item Q1. Does introducing data from multiple modalities into FL improve its
    performance?
    \item Q2. Does a classifier trained on labelled data from one modality work
    on testing data from other modalities?
    \item Q3. Does learning from both unimodal and multimodal clients provide
    better performance than only learning from multimodal clients?
\end{itemize}

\label{sev:eval}
\subsection{Datasets}
As human activity recognition (HAR) is a domain that often relies on multimodal
data, we used three HAR datasets that contain IoT data from different modalities
in our experiments. Table~\ref{tab:datasets} shows the modalities, $X$ sizes,
$h$ sizes, and the number of classes in the datasets.

\begin{table}[t]
	\caption{USED MULTIMODAL DATASETS}
	\label{tab:datasets}
	\centering
	\begin{tabular}{ccccc}
	\toprule
	\textbf{Dataset} & \textbf{Modality} & \textbf{$X$ size} & \textbf{$h$ size} & \textbf{Classes}\\
	\midrule
	Opp& 
	\begin{tabular}{@{}c@{}}Acce\\Gyro \end{tabular} & 
	\begin{tabular}{@{}c@{}}24\\15\end{tabular}&
	10 & 18\\
	\midrule
	mHealth& 
	\begin{tabular}{@{}c@{}}Acce\\Gyro\\Mag\end{tabular} & 
	\begin{tabular}{@{}c@{}}9\\6\\6\end{tabular}&
	4 & 13\\
	\midrule
	UR Fall& 
	\begin{tabular}{@{}c@{}}Acce\\RGB\\Depth\end{tabular} & 
	\begin{tabular}{@{}c@{}}3\\512\\8\end{tabular}&
	2,4 & 
	3\\
	\bottomrule
	\end{tabular}
\end{table}

\subsubsection{Different sensory modalities}
The Opportunity (Opp) challenge dataset~\cite{Chavarriaga2013} contains 18
short-term and non-repeated kitchen activities including
\emph{opening \& closing doors, fridges, dishwashers, and drawers, cleaning tables,
drinking from cups, toggling switches}, and \emph{null activities}. Its multimodal data
are measured by on-body sensors including accelerometers, gyroscopes, and
magnetic sensors. We use the accelerometer data (Acce) measured
in $milligrams$ and gyroscope data (Gyro) measured in $degrees/s$  as
the two modalities in our experiments. Following the experimental setup used by
Hammerla~\etal~\cite{Hammerla2016}, we use the runs ADL4 and ADL5 of subjects 2
and 3 as testing data ($118k$ samples) and the remaining runs
(except for ADL2 of subject 1) as training data ($525k$ samples).
For \textit{NaN} data in a sequence, we use their previous value in the sequence
to replace them~\cite{Chavarriaga2013}. As the training data are from 15 runs,
when generating local data for a client, the size of the randomly sampled
sequence is $1/15$ of the training data.

The mHealth dataset~\cite{O.2014} contains 13 daily living and exercise
activities including \emph{standing still, sitting \& relaxing, lying down,
walking, climbing stairs, waist bending forward, frontal elevation of arms,
knees bending, cycling, jogging, running, jumping front \& back}, and \emph{null
activities}. The activities are measured by multimodal on-body sensors including
accelerometers, ECG sensors, gyroscopes, and magnetometers. We use the
accelerometer data (Acce) measured in $meters/s^2$ , gyroscope data (Gyro)
measured in $degrees/s$, and magnetometer data (Mag) measured as local magnetic
field in our experiments and test the combinations of each two of them. For each
replicate of our simulations, we use the Leave-One-Subject-Out method to
randomly choose one participant and use her data as testing data. The other 9
participants' data are used as training data. The average number of samples from
a participant is $122\pm18k$ ($mean\pm std$). The size of the randomly sampled
sequence for a client is $1/9$ of the training data.

\subsubsection{Sensory-Visual modalities}
The UR Fall Detection dataset~\cite{Kwolek2014} contains 70 video clips recorded
by a RGB camera (RGB) and a depth camera (Depth) of human activities including
\emph{not lying, lying on the ground}, and \emph{temporary poses}.
Each video frame is labelled and paired with sensory data from accelerometers
(Acce) measured in $grams$. We use this dataset for our
experiments on sensory-visual and visual-visual modality combinations. For the
modality RGB, similar to the work by Srivastava~\etal~\cite{Srivastava2015}, we
use a pre-trained ResNet-18~\cite{He2016} to convert each frame into a feature
map. For the modality Depth, we use the extracted features
including \emph{HeightWidthRatio, MajorMinorRatio,
BoundingBoxOccupancy, MaxStdXZ, HHmaxRatio, Height, Distance}, and \emph{P40Ratio},
which are provided in the dataset. The size of $h$ is 2 with Acce and is 4
without it. For each replicate of our simulations, we randomly sample $1/10$
data (\ie, 7 video clips) as testing data and use the rest as training data.
The average number of frames in a video clip is $164\pm82$ ($mean\pm std$).
From the training data, the size of a randomly sampled sequence for a client is
$1/9$ of the training data.

\subsection{Simulation setup}
In each replicate of our simulation, the server conducts at most 100
communication rounds with the clients and selects 10\% clients for local
training (2 epochs with a 0.01 or a 0.001 learning rate, whichever provides
better performance) in each round, after which the cloud training (5 epochs with
a 0.001 learning rate) is conducted. The labelled dataset on the server is
randomly sampled from the training dataset and its size is the same as the size
of a client's local data. For DCCAE, we set $\lambda=0.01$ as suggested by
Wang~\etal~\cite{Wang2015}. For the multimodal weight parameter $\alpha$, we
tested $\{1, 2, 10, 50, 100, 500\}$ and found that $\alpha=100$ provides the
best performance. For each individual simulation setup, we use different random
seeds to run 64 replicates.

\subsubsection{Baselines}
To answer Q1, we consider a system in which clients have multimodal data and a
server has two labelled unimodal datasets. Without multimodal representation
learning, a baseline scheme can only use data from one modality, which we refer
to as \textbf{UmFL} (30 unimodal clients, 1 label modality). Comparing UmFL
with our multimodal scheme (30 multimodal clients, 2 label modalities) will
reveal whether introducing more modalities in FL improves its performance. We
test both of them on the data from the modality of UmFL.

To answer Q2, we consider a system wherein clients have multimodal data and a
server has a labelled dataset from one modality. A baseline scheme trains a
global unimodal autoencoder for each modality with the same size of $h$. The
classifier of the baseline is trained on the labelled data from one modality
with the help from the autoencoder on that modality. We directly test the
classifier on data from the other modality, since the sizes of $h$ from two
modalities are the same. This baseline does not use the
alignment information to do any multimodal local training. It is for the
ablation study on the multimodal local training and multimodal FedAvg
component. We refer to this baseline as \textbf{Abl} (30 unimodal clients for
each modality, 1 label modality).  Comparing Abl with our scheme (30 multimodal
clients, 1 label modality) will indicate whether the multimodal component brings
any improvement to the performance.

To answer Q3, we consider a system that has both unimodal clients and multimodal
clients. The server in the system has a labelled dataset from one modality. A
baseline scheme only chooses multimodal clients (30 clients) to update the
global autoencoders. Comparing it with other schemes that use both multimodal
and unimodal clients for local update will show whether our proposed Mm-FedAvg
improves the performance of the system.

\begin{figure*}[t]
    \centering
   \subfloat[][Opp (Acce \& Gyro)]{\includegraphics[width=.30\linewidth]{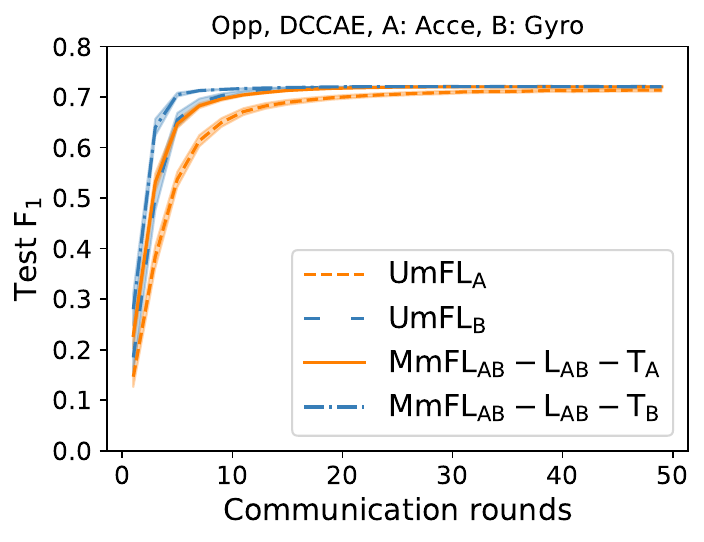}\label{sub_fig:single-multi-opp}} 
   \subfloat[][mHealth (Acce \& Gyro)]{\includegraphics[width=.30\linewidth]{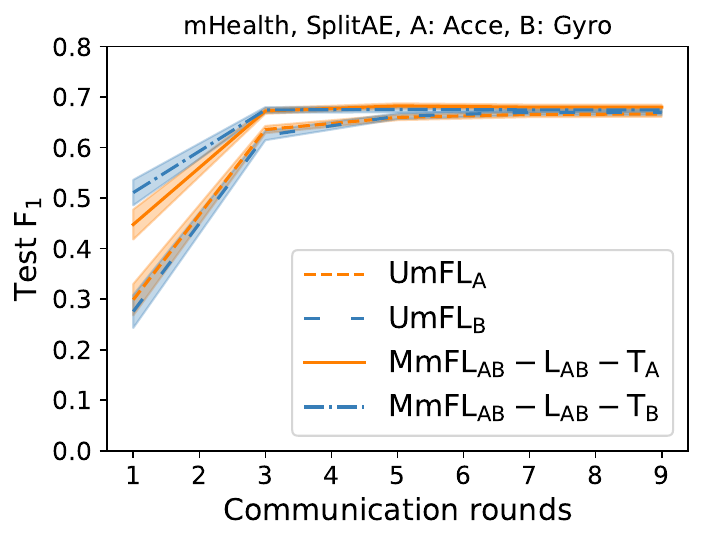}\label{sub_fig:single-multi-mhealth-acce-gyro}} 
   \subfloat[][mHealth (Acce \& Mag)]{\includegraphics[width=.30\linewidth]{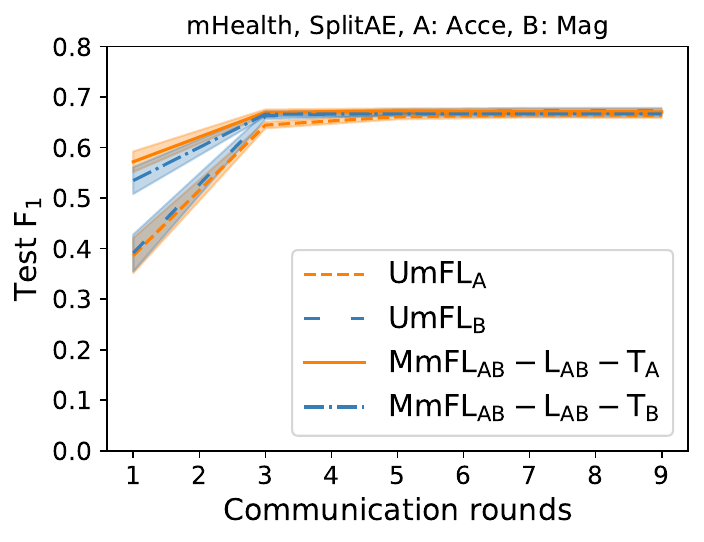}\label{sub_fig:single-multi-mhealth-acce-mage}} \\
   \subfloat[][mHealth (Gyro \& Mag)]{\includegraphics[width=.30\linewidth]{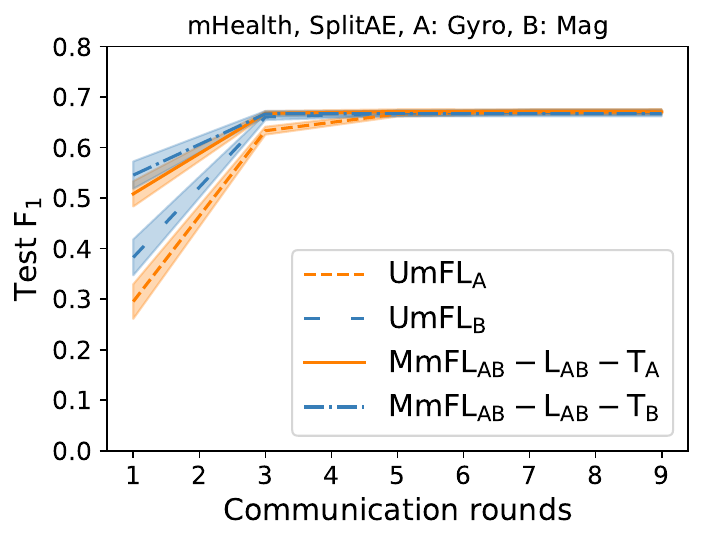}\label{sub_fig:single-multi-mhealth-gyro-mage}} 
   \subfloat[][UR Fall (Acce \& Depth)]{\includegraphics[width=.30\linewidth]{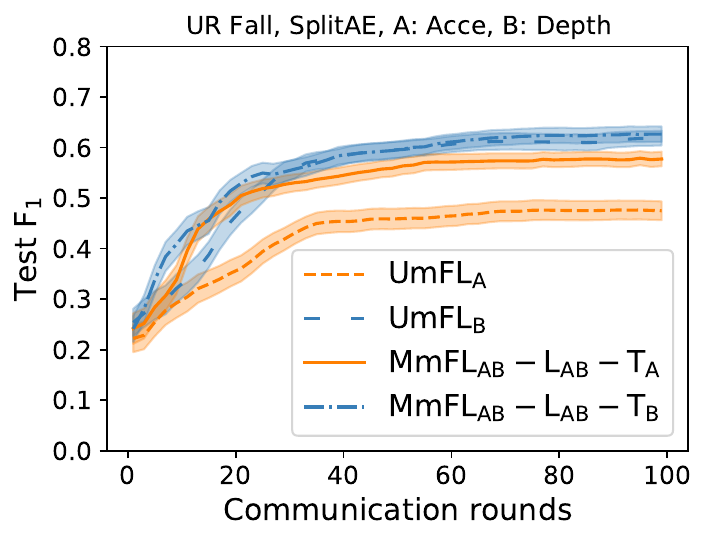}\label{sub_fig:single-multi-urfall-acce-depth}}
   \subfloat[][UR Fall (RGB \& Depth)]{\includegraphics[width=.30\linewidth]{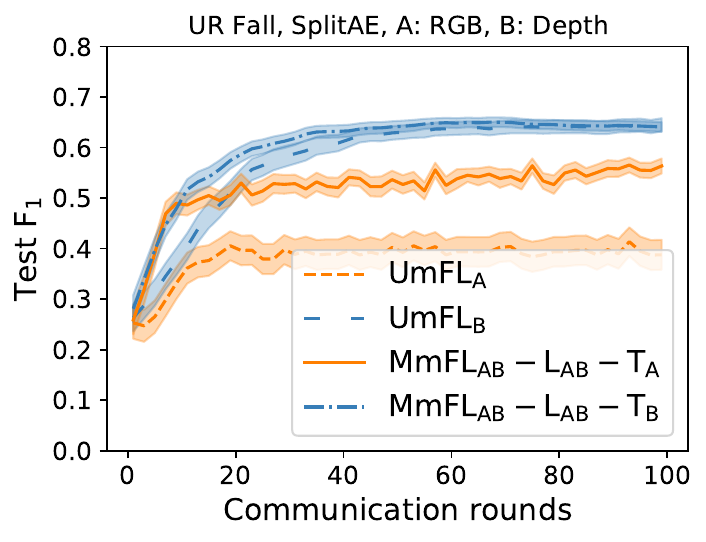}\label{sub_fig:single-multi-urfall-rgb-depth}}
    \caption{Comparison between UmFL and MmFL. MmFL schemes have higher or same
    level of converged $F_1$ scores on UR Fall datasets than UmFL schemes do. On
    all three datasets, MmFL converges faster than UmFL does.}
    \label{fig:single-multi-comparison}
\end{figure*}

\subsubsection{Models}
We implement all the deep learning components through the PyTorch
library~\cite{pytorch}. For training autoencoders on time-series data, we use
long short-term memory (LSTM)~\cite{lstm} autoencoders~\cite{Srivastava2015} in
our experiments for local training and use the bagging strategy~\cite{Guan2017}
to train our models with random batch sizes and sequence lengths. An LSTM
autoencoder takes a time-series sequence (\eg, sensory data, video frames) as
its input. The hidden states generated by the LSTM encoder unit are used as the
hidden representations of the input samples in the sequence. On the server side,
we use a simple classifier that has one multilayer perceptron (MLP) layer
connected to one LogSoftmax layer as the model for supervised learning. On the
mHealth dataset, we introduce a Dropout layer (rate=0.5) before the MLP layer of
the classifier to prevent overfitting.

\subsection{Metrics}
We test the classifier on the server against a labelled testing dataset. We use
a sliding time window with length of 2,000 to extract time-series sequences
(without overlap) from the testing dataset. We use the encoder of $w^{a_{g}}$
for the modality of the testing data to convert the sequences into
representations and test them on the classifier $w^{s}$. We calculate the
$F_1$ score of each class within a sequence as:

\begin{equation}
F_1 = \frac{2*TP}{2*TP+FP+FN}
\end{equation}

TP, FP, and FN are the numbers of true positive, false positive,
and false negative classification results, respectively. The weighted average
$F_1$ score of all classes within the sequence (with the number of ground truth
samples of a class being its weight) is the $F_1$ score on the sequence. And the
average $F_1$ score of all sequences is the $F_1$ score of the classifier. We
evaluate the $F_1$ score of the classifier every other communication round until
it converges and calculate its average value and standard error from 64
replicates.  On each dataset, we evaluate both SplitAE and DCCAE and keep the
one that has better $F_1$ scores.

\begin{figure*}[t]
    \centering
   \subfloat[][Opp (Acce \& Gyro)]{\includegraphics[width=.5\linewidth]{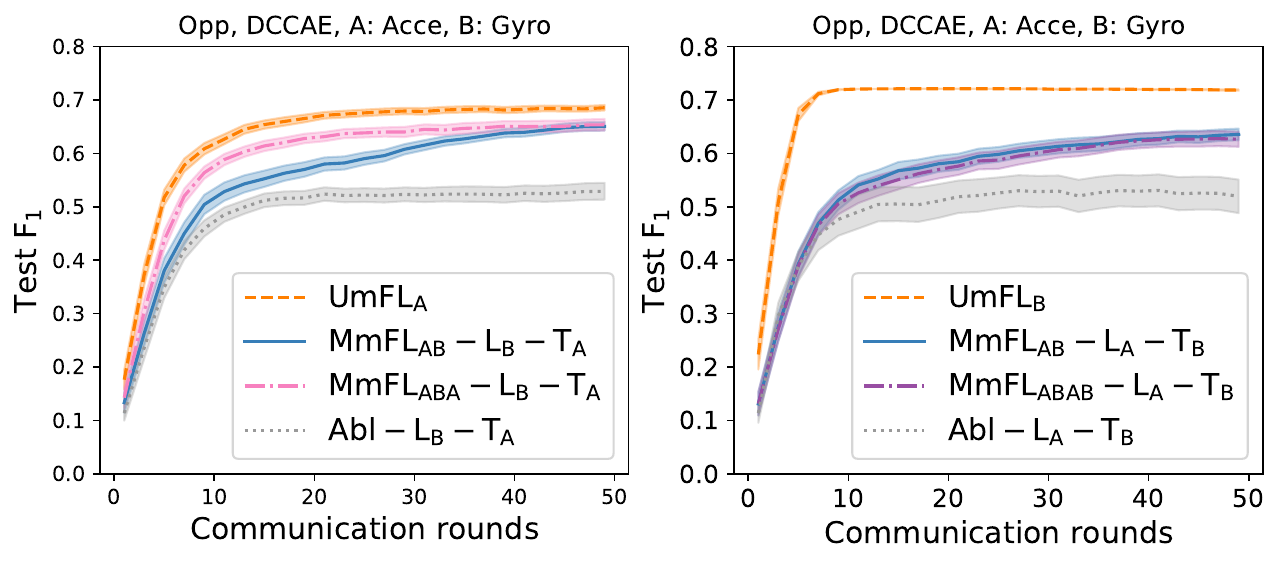}\label{sub_fig:cross-opp}}
   \subfloat[][mHealth (Acce \& Gyro)]{\includegraphics[width=.5\linewidth]{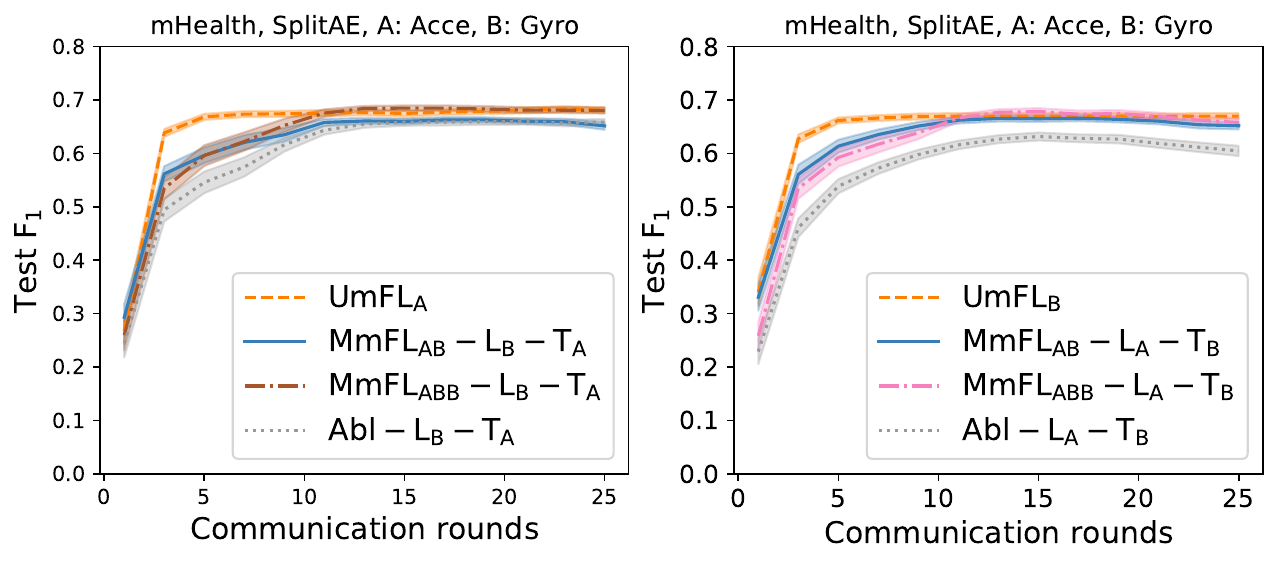}\label{sub_fig:cross-mhealth-acce-gyro}}\\ 
   \subfloat[][mHealth (Acce \& Mag)]{\includegraphics[width=.5\linewidth]{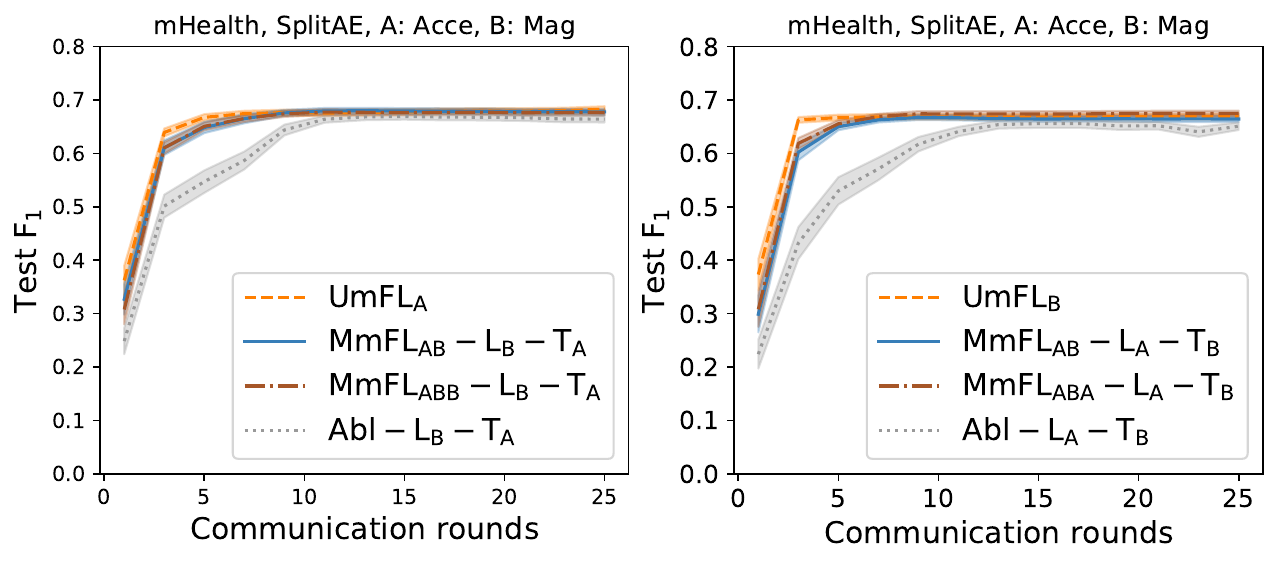}\label{sub_fig:cross-mhealth-acce-mage}}
   \subfloat[][mHealth (Gyro \& Mag)]{\includegraphics[width=.5\linewidth]{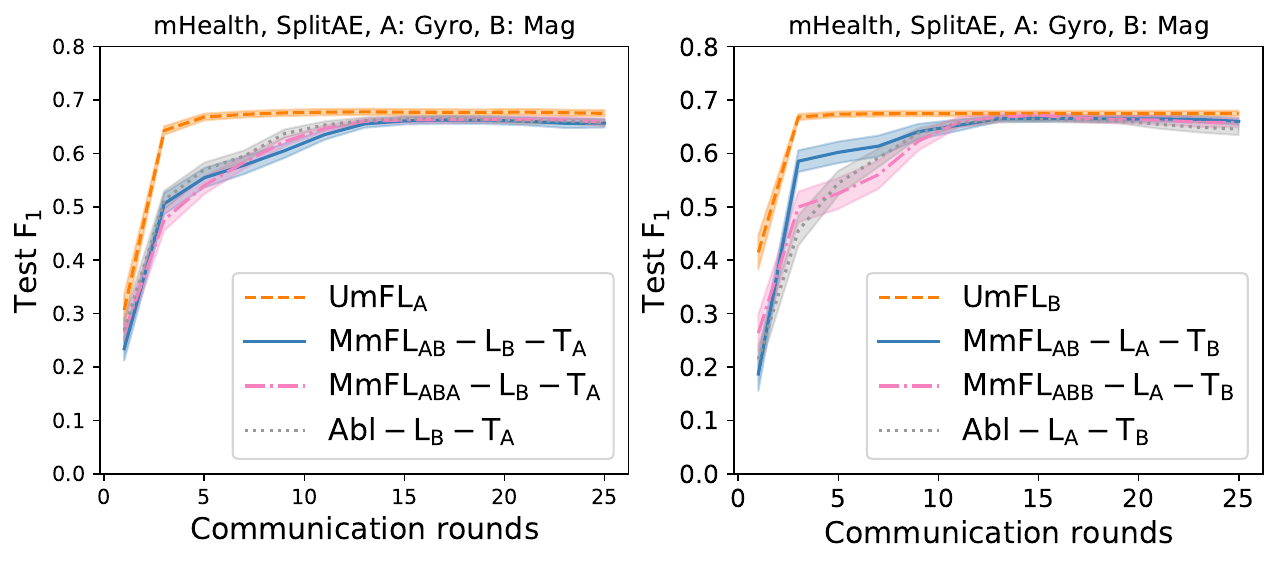}\label{sub_fig:cross-mhealth-gyro-mage}}\\
   \subfloat[][UR Fall (Acce \& Depth)]{\includegraphics[width=.5\linewidth]{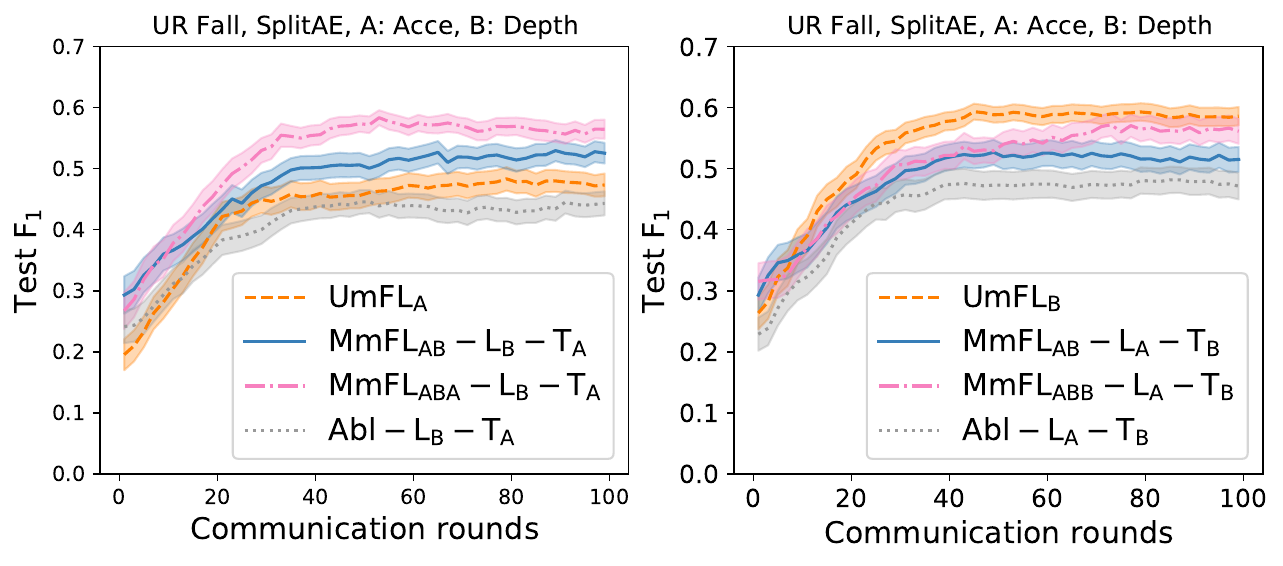}\label{sub_fig:cross-urfall-acce-depth}}
   \subfloat[][UR Fall (RGB \& Depth)]{\includegraphics[width=.5\linewidth]{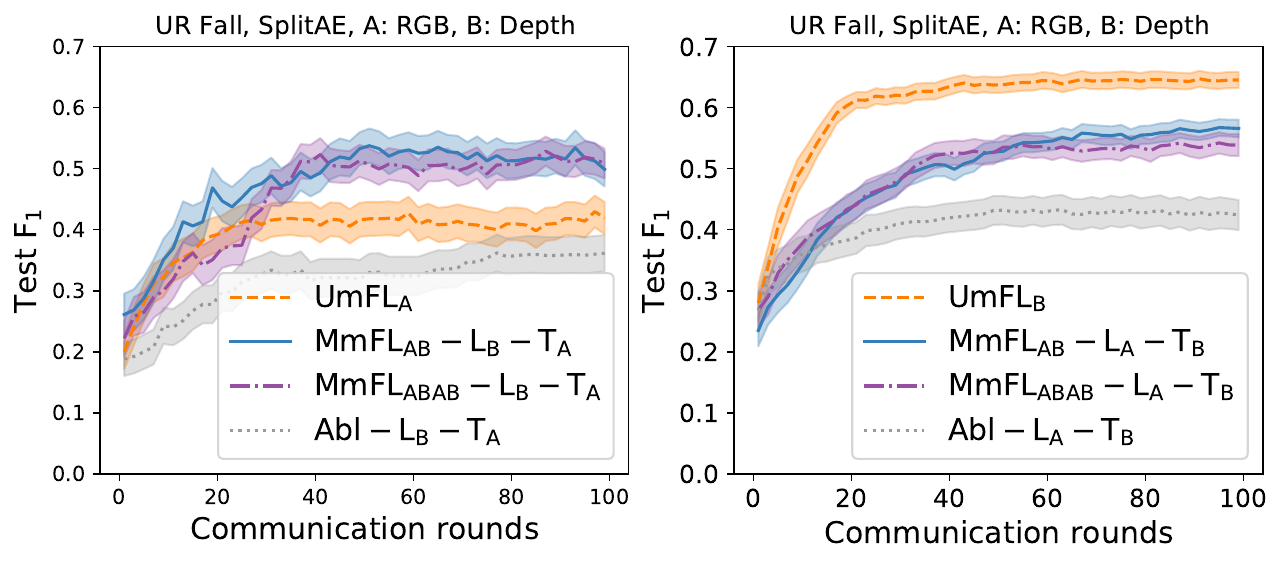}\label{sub_fig:cross-urfall-rgb-depth}}
    \caption{$F_1$ scores of MmFL with labelled data from one modality (\eg,
    L\textsubscript{B}) and test data from the other modality (\eg,
    T\textsubscript{A}). MmFL schemes achieve higher converged $F_1$ scores or
    faster convergence than baselines (\ie, Abl schemes) in most cases.
    Combining contributions from both unimodal and multimodal clients (\eg,
    MmFL\textsubscript{ABA}) can further improve the $F_1$ scores.}
    \label{fig:cross-comparison}
\end{figure*}

\section{Results}
\label{sec:results}
We find that by using data from multiple modalities, the $F_1$ score of the
classifier is higher than that by using data from one single modality. With the
help of multimodal representations, the classifier trained on labelled data from
one modality can be used on the data from another modality and achieve
acceptable $F_1$ scores. In addition, combining local autoencoders from both
unimodal and multimodal clients can achieve higher $F_1$ scores than only using
multimodal clients.

\subsection{Multimodal data improve $F_1$ scores}
\label{subsec:single-multi-modality}
On the Opp dataset, as shown in Fig.~\ref{sub_fig:single-multi-opp}, the $F_1$
scores of multimodal schemes (MmFL) that are trained on labelled datasets from
two modalities (L\textsubscript{AB}) converge faster than UmFL\textsubscript{A}
and UmFL\textsubscript{B} do when being tested on each modality
(T\textsubscript{A} and T\textsubscript{B}). Although the converged $F_1$ scores
are the same for both UmFL and MmFL, using multimodal data speeds up the
convergence.

On the mHealth dataset
(Fig.~\ref{sub_fig:single-multi-mhealth-acce-gyro}--~\ref{sub_fig:single-multi-mhealth-gyro-mage}),
the results on three modality combinations show similar trends. On each testing
modality, the converged $F_1$ scores of MmFL schemes are similar to those of
their unimodal counterparts. However, the $F_1$ scores of MmFL schemes converge
faster than UmFL schemes do.

On the UR Fall dataset, the sizes of $X$ from Acce and RGB are 3 and 512,
respectively. Thus $h=2$ is the largest representation size that we can use for
the modality combination Acce \& RGB and it is not large enough to encode useful
representations from RGB data. Therefore we only show the results from the other
two modality combinations (Fig.~\ref{sub_fig:single-multi-urfall-acce-depth} \&
~\ref{sub_fig:single-multi-urfall-rgb-depth}). The $F_1$ scores of MmFL schemes
are higher than those of UmFL schemes when the schemes are tested against Acce
data or RGB data. When being tested against Depth data, MmFL schemes converge
faster than UmFL schemes do. Even the modalities of data in UR Fall are more
heterogeneous (\ie, sensory \& visual) than those in Opp or mHealth (\ie,
sensory \& sensory), multimodal FL can still align their representations,
thereby introducing more data to improve the $F_1$ score of the FL system.

Similar to the results of existing studies on centralized ML systems, our
results demonstrate that, in FL systems, combining different modalities through
multimodal representation learning can achieve higher $F_1$ scores or faster
convergence than only using unimodal data. Compared with existing work using
early fusion~\cite{Liang2020}, the labelled data source on the server in our
framework does not have to be aligned multimodal data. It can be individual
unimodal datasets that are collected separately. This suggests that we can scale
up FL systems across different modalities by utilizing the alignment information
contained in local data on multimodal clients.

\subsection{Labels can be used across modalities}
\label{subsec:cross-modality}
To answer Q2, we use labelled data from one modality for supervised learning on
the server and test the trained classifier on the other modality that does not
have any labels in the system. Fig.~\ref{fig:cross-comparison} shows the
$F_1$ scores of MmFL with different modalities for labelled data (\eg,
L\textsubscript{B}) and testing data (\eg, T\textsubscript{A}), in comparison
with a baseline scheme (Abl) for the ablation study and a unimodal scheme for
the modality of the testing data (\eg, UmFL\textsubscript{A}).

On the Opp dataset with DCCAE (Fig.~\ref{sub_fig:cross-opp}), using only
multimodal clients (\ie, MmFL\textsubscript{AB}) achieves higher converged $F_1$
scores than baseline schemes do, which means that the multimodal representation
learning on clients indeed aligns two modalities. When training classifiers on
labelled Gyro data and testing them on Acce data (\ie,
MmFL\textsubscript{AB}-L\textsubscript{B}-T\textsubscript{A}),
the $F_1$ score is close to that of a unimodal scheme using Acce data (\ie,
UmFL\textsubscript{A}), which demands labelled Acce data on the server.

On the mHealth dataset
(Fig.~\ref{sub_fig:cross-mhealth-acce-gyro}--\ref{sub_fig:cross-mhealth-gyro-mage}),
the converged $F_1$ score of baseline schemes and unimodal schemes is close to
each other. This means that the different modalities may be correlated even without
being aligned (similar to the findings reported by
Malekzadeh~\etal~\cite{malekzadeh2020dana}). This might be due to the fact that
except for 1 accelerometer on the chest, 6 sensors for different modalities in
the mHealth dataset were attached to 2 body parts (\eg, left-ankle and
right-lower-arm). Thus the readings of different modalities from the same body
part might be correlated. MmFL\textsubscript{AB} schemes still improve the
converged $F_1$ scores compared to Abl schemes and have faster convergence in
two modality combinations (\ie, Acce \& Gyro, Acce \& Mag).

On the UR Fall dataset
(Fig.~\ref{sub_fig:cross-urfall-acce-depth}--\ref{sub_fig:cross-urfall-rgb-depth}),
MmFL\textsubscript{AB} schemes have higher $F_1$ scores than baselines do. It is
worth to note that, when using labelled Depth data (\ie, L\textsubscript{B}), the test
$F_1$ scores on Acce and RGB data (\ie,
MmFL\textsubscript{AB}-L\textsubscript{B}-T\textsubscript{A} schemes in
Fig.~\ref{sub_fig:cross-urfall-acce-depth} \&
~\ref{sub_fig:cross-urfall-rgb-depth}) are even higher than those when using labelled
data from these two testing modalities (\ie, UmFL\textsubscript{A}). In
Sec.~\ref{subsec:single-multi-modality}, results in
Fig.~\ref{sub_fig:single-multi-urfall-acce-depth} \&
~\ref{sub_fig:single-multi-urfall-rgb-depth} show that the unimodal schemes
using Depth data have higher $F_1$ scores than those using Acce or RGB data.
Therefore, for MmFL with SplitAE, using labelled Depth data for the supervised
learning on the server leads to higher $F_1$ scores than those using Acce or RGB
data's own labels.

Our results show that, with the help of multimodal representation learning on FL
clients, we can use the trained global autoencoder to share the label
information from one modality to other modalities by mapping them into
shared or related representations. The test $F_1$ scores on the other
modalities can be close to or even better than those of unimodal FL schemes
using labels from the modalities. This allows us to scale up FL systems even
with limited source of unimodal labelled data. In addition, we can potentially
improve the testing performance of a modality by aligning it with other
modalities that have labels, instead of directly mapping it to labels.

\subsection{Training on mixed clients} 
To understand how mixed clients with different device setups (\ie,
unimodal clients and multimodal clients), which is a more realistic scenario
for FL systems, affect the $F_1$ scores, for each MmFL\textsubscript{AB} scheme
with 30 multimodal clients, we run one mixed-client scheme that has 10 more
clients for modality $A$ (\ie, MmFL\textsubscript{ABA}), one that has 10 more
clients for modality $B$ (\ie, MmFL\textsubscript{ABB}), and one that has 10
more clients for each modality (\ie, MmFL\textsubscript{ABAB}). We compare them
and keep the one that has the highest $F_1$ scores.

In Fig.~\ref{sub_fig:cross-opp}, the
MmFL\textsubscript{ABA}-L\textsubscript{B}-T\textsubscript{A} scheme on the Opp
dataset further speeds up the convergence of test $F_1$ scores compared to
MmFL\textsubscript{AB}, which means that combining contributions from both
unimodal and multimodal clients by using Mm-FedAvg is better than using only
multimodal clients. On the mHealth dataset
(Fig.~\ref{sub_fig:cross-mhealth-acce-gyro} \&
~\ref{sub_fig:cross-mhealth-acce-mage}), the mixed-client schemes slightly
improve the test $F_1$ scores in two experiments. Similarly, on the UR Fall
dataset (Fig.~\ref{sub_fig:cross-urfall-acce-depth}), MmFL\textsubscript{ABA}
and MmFL\textsubscript{ABB} schemes show improved $F_1$ scores in the
experiments of the Acce \& Depth combination.

The results indicate that using Mm-FedAvg to combine models from both multimodal
(with higher weights) and unimodal clients can provide higher $F_1$ scores or
faster convergence than only using multimodal clients. Thus, when there are a
limited number of multimodal clients in a mixed-client FL system, we can utilize
unimodal clients to boost the local training.

\section{Discussions}
In this paper, we have proposed a multimodal FL framework on IoT data. We now
discuss how the framework can be used in real-world FL systems and what
potential research topics are in the space of multimodal FL.

\subsection{Heterogeneity beyond data distributions}
Training in FL is mainly conducted on clients. In a real-world FL system, each
client's local data are generated on an individual level rather than a
population level, which means that heterogeneity between clients is commonplace.
Some heterogeneity such as data distributions has been well studied and solving
it can help keep the performance of FL systems stable across different
clients. Other heterogeneity, such as data modalities, is also an important
issues in implementing FL systems. As shown in our results, solving such
heterogeneity can make FL systems scalable across different modalities, thereby
increasing the amount of available data. In an FL system using IoT devices, it
is difficult to force all clients to deploy devices that have the same data
modality, because users may have different budgets for devices or privacy
concerns on the devices installed in their homes. Therefore, multimodal FL plays
an important role in realizing those promised FL systems that aim to work with
hundreds of thousands of clients. In this paper, we focused on
the modality heterogeneity issue and the other types of heterogeneity are out of
our scope, which is the limitation of this paper. For future research, we plan
to investigate how multimodal FL performs with the influence from the other
types of heterogeneity in aspects such as data distributions and DNN model
structures.

\subsection{Sharing label information across modalities}
The lack of labelled data on FL clients has recently motivated researchers to
design semi-supervised FL systems. In many cases, only the service provider (\ie,
the FL server) has the ability and expertise to provide labelled data. The
existing research on semi-supervised FL assumes that the labelled data on the
server and the local data on clients are from the same modality. In this paper,
we have shown that our framework allows label information from one modality to
be used by other modalities. This can potentially contribute to reducing the
cost of data annotation on the server when implementing real-world
semi-supervised FL systems. Some modalities (\eg, sensory data) may not be easy
to directly annotate on. However, by using the matching information on FL
clients, we can align these modalities with other modalities that are easy to
acquire annotations (\eg, visual data) on the server. By this means, we can
enable clients from all modalities in the system to utilize the label
information through multimodal representations. It may also allow us to deploy
fewer privacy-intrusive devices (\eg, cameras) in people's homes since we only
need some clients to have multimodal data for alignment.

\subsection{Utilizing mixed FL clients}
One of our contributions in this paper is the Mm-FedAvg algorithm that combines
locally updated autoencoders from both unimodal and multimodal clients. By
giving multimodal clients more weights, combining contributions from mixed
clients has higher $F_1$ scores than only using multimodal clients. Thus
only a part of the clients in the system needs to be multimodal clients.
Currently, all the multimodal clients in the framework use the same type of
autoencoder (\ie, either all SplitAE or all DCCAE) and the unimodal clients' can
directly update a part of the autoencoders. In reality, this assumption may
need to be changed due to different local data distributions or computational
capabilities. Therefore, we suggest that more flexible multimodal averaging
algorithms using techniques such as knowledge
distillation~\cite{lin2020ensemble} should be investigated. It would allow FL
systems to use different local autoencoders for multimodal representation
learning. In addition, mechanisms that can evaluate the quality of models
trained on different data modalities and can dynamically adjust the weights of
multimodal clients are necessary, which will allow us to optimise the combined
contributions.

\section{Conclusions}
As a new system paradigm, federated learning (FL) has shown great potentials to
realize deep learning systems in the real world and protect the privacy of data
subjects at the same time. In this paper, we propose a multimodal and
semi-supervised framework that enables FL systems to work with clients that have
local data from different modalities and clients with different device setups (\ie,
unimodal clients and multimodal clients). Our experimental results demonstrate
that introducing data from multiple modalities into FL systems can improve their
classification $F_1$ scores. In addition, it allows us to apply models trained
on labelled data from one modality to testing data from other modalities and
achieve decent $F_1$ scores. It only requires a part of the clients to be
multimodal in order to align different modalities. We believe that our
contributions can help machine-learning system designers who want to implement
FL in complex real-world scenarios such as IoT environments, wherein data are
generated from different modalities. For future research, we plan to investigate
broader applications of our framework in domains apart from multimodal human
activity recognition.

\section*{Acknowledgement}
This work was supported by the UK Dementia Research Institute.

\bibliographystyle{IEEETran}
\balance
\bibliography{paper}
\end{document}